\newif\ifarxiv
\arxivtrue % comment out to exit arxiv mode
\newif\ifNeurocomputing
\ifarxiv
    \documentclass{article}
\fi
\ifNeurocomputing
    \documentclass[preprint, number, sort&compress]{elsarticle}
\fi

\usepackage{lineno,hyperref,float}
\usepackage{pgfplots, pgfplotstable}
\usepackage{multirow}
\usepackage{multicol}
\usepackage{xcolor}
\usepackage{ifthen}

\ifarxiv
    \usepackage[affil-it]{authblk}
\fi

\newcommand\revision[1]{\textcolor{black}{#1}}

\modulolinenumbers[5]

\ifNeurocomputing
    \journal{Journal of Neurocomputing}
\fi

\ifarxiv
\fi
\begin{document}

\ifNeurocomputing
    \begin{frontmatter}
\fi

\title{Encrypted Internet Traffic Classification Using a Supervised Spiking Neural Network}

\ifNeurocomputing
    \author[Sharif]{Ali Rasteh}
    \author[Toulouse]{Florian Delpech}
    \author[Toulouse]{Carlos Aguilar-Melchor}
    \author[CNRS]{Romain Zimmer}
    \author[Sharif]{Saeed Bagheri Shouraki}
    \author[CNRS]{Timothée Masquelier\corref{mycorrespondingauthor}}
    \cortext[mycorrespondingauthor]{Corresponding author}
    \ead{timothee.masquelier@cnrs.fr}
    
    \address[Toulouse]{Institut Supérieur de l’Aéronautique et de l’Espace (ISAE-SUPAERO), University of Toulouse, France}
    \address[CNRS]{Cerco UMR 5549, CNRS, Université Toulouse 3, Toulouse, France}
    \address[Sharif]{Artificial Creatures Laboratory, Electrical Engineering Department, Sharif University of Technology, Tehran, Iran}
\fi

\ifarxiv
    % \author{Ali Rasteh$^{3}$\\
    % Florian Delpech$^{1}$\\
    % Carlos Aguilar-Melchor$^{1}$\\
    % Romain Zimmer$^{2}$\\
    % Saeed Bagheri Shouraki$^{3}$\\
    % Timothée Masquelier$^{2}$}
    
    \author{Ali Rasteh$^{3,}$%
	    \thanks{e-mail: \texttt{ali.rasteh@ee.sharif.edu}}}
    \author{Florian Delpech$^{1,}$}
    \author{Carlos Aguilar-Melchor$^{1,}$}
    \author{Romain Zimmer$^{2,}$}
    \author{Saeed Bagheri Shouraki$^{3,}$}
    \author{Timothée Masquelier$^{2}$}

    % \newenvironment{affiliations}{%
    %     \setcounter{enumi}{1}%
    %     \setlength{\parindent}{0in}%
    %     \slshape\sloppy%
    %     \begin{list}{\upshape$^{\arabic{enumi}}$}{%
    %         \usecounter{enumi}%
    %         \setlength{\leftmargin}{0in}%
    %         \setlength{\topsep}{0in}%
    %         \setlength{\labelsep}{0in}%
    %         \setlength{\labelwidth}{0in}%
    %         \setlength{\listparindent}{0in}%
    %         \setlength{\itemsep}{0ex}%
    %         \setlength{\parsep}{0in}%
    %         }
    %     }{\end{list}\par\vspace{12pt}}
        
    % \begin{affiliations}
    % \item {Institut Supérieur de l’Aéronautique et de l’Espace (ISAE-SUPAERO), University of Toulouse, France
    % }
    % \item {Cerco UMR 5549, CNRS, Université Toulouse 3, Toulouse, France
    % }
    % \item {Artificial Creatures Laboratory, Electrical Engineering Department, Sharif University of Technology, Tehran, Iran
    % }
    % \end{affiliations}
    
    \affil{\normalsize $^1$Institut Supérieur de l’Aéronautique et de l’Espace (ISAE-SUPAERO), University of Toulouse, France}
    \affil{$^2$Cerco UMR 5549, CNRS, Université Toulouse 3, Toulouse, France}
    \affil{$^3$Artificial Creatures Laboratory, Electrical Engineering Department, Sharif University of Technology, Tehran, Iran}
    
    \maketitle
\fi
% ===================================================

\begin{abstract}
Internet traffic recognition is essential for access providers since it helps them define adapted priorities in order to enhance user experience, e.g., a high priority for an audio conference and a low priority for a file transfer. As internet traffic becomes increasingly encrypted, the main classic traffic recognition technique, payload inspection, is rendered ineffective. Hence this paper uses machine learning techniques looking only at packet size and time of arrival. For the first time, Spiking neural networks (SNNs), which are inspired by biological neurons, were used for this task for two reasons. Firstly, they can recognize time-related data packet features. Secondly, they can be implemented efficiently on neuromorphic hardware. Here we used a simple feedforward SNN, with only one fully connected hidden layer, and trained in a supervised manner using the new method known as Surrogate Gradient Learning. Surprisingly, such a simple SNN reached an accuracy of 95.9\% on ISCX datasets, outperforming previous approaches. Besides better accuracy, there is also a significant improvement in simplicity: input size, the number of neurons, trainable parameters are all reduced by one to four orders of magnitude. Next, we analyzed the reasons for this good performance. It turns out that, beyond spatial (i.e., packet size) features, the SNN also exploits temporal ones, mainly the nearly synchronous (i.e., within a 200ms range) arrival times of packets with specific sizes. Taken together, these results show that SNNs are an excellent fit for encrypted internet traffic classification: they can be more accurate than conventional artificial neural networks (ANN), and they could be implemented efficiently on low-power embedded systems.
\end{abstract}

% ===================================================

\ifNeurocomputing
    \begin{keyword}
    \texttt Spiking Neural Network \sep Surrogate Gradient Learning \sep Internet traffic classification
    \end{keyword}
\fi

\ifNeurocomputing
    \end{frontmatter}
\fi

%\linenumbers

% ===================================================

\section{Introduction}
\label{sec:Introduction}

Today, more than half of the global population uses the Internet network. The amount of data exchanged through it increases each year endlessly, giving people more access to information. Whether sending emails, downloading or uploading files, watching videos on streaming platforms, chatting on social media, etc., all these operations involve devices interacting with computer servers located around the world. To improve the integrity of traffic flows, a protocol stack named TCP-IP is used (see e.g.~\cite{peterson2007computer} chap. 5). When data are exchanged via this protocol stack, they are broken down into different packets. Traffic classification consists of associating traffic flows comprised of data packets, with the categories corresponding to different use cases (e.g., email, chat, file transfer, etc.). 

Increasing demands to protect privacy have led to the development of methods to encrypt traffic, and of different ways to navigate on the Internet anonymously, such as The Onion Router (Tor)\cite{dingledine2004tor} or Virtual Private Networks (VPN) (see e.g.~\cite{scott1999virtual}). These techniques encrypt the whole original packets (payload and headers) into a new encrypted payload, adding a new header to ensure traffic handling in the VPN or Tor network. Thus neither the original payload nor header are available for inspection, and the only features that can be used for classification are the encrypted size of the original packet and the time at which the packet is captured. With only these features, packet inspection is not possible anymore, and statistical or machine learning techniques must be used for traffic classification. The state-of-the-art results for such an approach to traffic classification are based on ANNs. 

There are different types of neural networks that are used in different contexts:
\begin{itemize}
    \item Convolutional Neural Networks (CNNs) \cite{LeCun2015} are mainly employed for image recognition and have a structure inspired by the primate's visual cortex.
    \item Recurrent Neural Networks (RNNs) \cite{Schmidhuber2015} are used for automatic speech or text recognition and take into account the temporal dimension by interconnecting neurons to form cycles in the networks.
    \item Spiking Neural Networks (SNNs) \cite{Maass1997} are aimed at imitating biological neural networks. A neuron fires an electrical impulse called "spike" when its membrane potential reaches a specific value, leading to a signal composed of spikes that propagate onto the subsequent neurons. They are very interesting for recognizing time-related patterns. \revision{Although SNNs are recurrent networks, their recursive architecture is the result of their neurons' structure, and in that way, they differ from RNNs, which have recurrent connections between different neurons. Also, as described, they are entirely based on biological observations in contrast to RNNs.}
\end{itemize}
The main challenge of SNNs is that their binary nature prevents the use of classic training methods like back-propagating gradient descent \cite{Neftci2019, wu2018spatio, wozniak2020deep}. To solve this, most current methods are rate-based, meaning that they only consider the number of spikes inside a temporal window, ignoring their times. This is a severe limitation as different spike trains can have the same number of spikes but present distinct temporal patterns. To solve this problem, different alternatives have been developed. \revision{For instance, Neftci et al. 2019 \cite{Neftci2019} proposed a surrogate gradient algorithm that provides spike compatible backpropagation and preserves the temporal patterns.} It consists of approximating the true gradient of the loss function while keeping good optimization properties. \revision{While training SNNs is hard; they can extract the temporal features of data naturally as the surrogate gradient approach has preserved it. This feature makes them well-suited for data with temporal properties, precisely like the case of internet traffic.}

\subsection{Contributions}
\label{sec:Contributions}

%\fbox{\parbox{.8\linewidth}{PERMANENT TODO before submission check that the results here and in the abstract are up to date.}}

This paper shows that SNNs lead to very promising results for traffic recognition of encrypted traffic. We beat the state of the art~\cite{Shapira2019} for almost every metric on accuracy, precision, or recall. The average accuracy for Tor traffic rises from 67.8\% to 98.6\%, and for unencrypted traffic from 85\% to 99.4\%. For VPN traffic, it rises from 98.4\% to 99.8\%. The number of errors is thus divided by 8 for VPN traffic and by 20 for Tor and unencrypted traffic.

It is important to note that for VPN traffic, the most challenging situation (browsing versus chat traffic) is not present as the dataset used by~\cite{Shapira2019} has no VPN browsing data. One versus all experiments show that SNNs lead to a very significant improvement with all the problematic categorizations such as File Transfer for Tor (increased from 55.8\% to 99.7\%), Browsing for Tor (from 90.6\% to 99.0\%), Browsing for unencrypted (90.6\% to 99.4\%) and Chat for Tor (89\% to 99.1\%).

These results are obtained with: simpler inputs (300x300 images versus 1500x1500 images), a simpler network (one hidden layer of 100 neurons versus a six hidden layer LeNet-5 CNN with one million neurons), fewer training parameters (31.5k versus 300k), and a higher testing data over training data ratio (20/80 versus 10/90).

Our approach is all but fancy; the merit is due to the superior capabilities of SNNs to detect and exploit time-related features. In Section \ref{sec:Results} we study this with some experiments that highlight that short-term synchronicity detection seems to be an essential feature for traffic categorization, and SNNs naturally excel at this.

\subsection{Outline}
\label{sec:Outline}
The rest of this paper is organized as follows. Section \ref{sec:Related works} gives a brief introduction to related works in the field of internet traffic classification. Section \ref{sec:Proposed model} gives a complete review of the proposed model for classification of internet traffic using an SNN and with the help of the supervised surrogate gradient method. In section \ref{sec:Experiments} and \ref{sec:Results} some experiments and corresponding results are given to investigate the power of the proposed model to classify internet traffic data. Conclusions and future work are presented in Section \ref{sec:Conclusion}.

% ===================================================

\section{Related works}
\label{sec:Related works}
Mainstream techniques for classifying unencrypted traffic, port-based classification, and deep packet inspection~\cite{dainotti2012issues}, and part of the alternative statistical and machine learning techniques, are unusable for encrypted traffic as they use features that are only available before encryption. Such related work will, therefore, not be described here. 

There are two usual classification problems: \textit{traffic categorization} (or characterization), which considers types of traffic such as VoIP, File Transfer, Video, Browsing, Chat, P2P, etc.; and \textit{application identification} which considers given applications such as Kazaa, BitTorrent, Google Hangouts Chat,  AIM, WhatsApp Web, etc. In this work, we focus on traffic categorization and leave application categorization for future work.

There is a rich literature on traffic categorization through machine learning methods (statistics based or neural network based). Multiple statistical approaches using features compatible with encryption (such as packet times and sizes) have led to scientific publications: fingerprinting \cite{crotti2007traffic, wang2010optimised, qin2015robust}, Bayes \cite{moore2013discriminators, moore2005internet, fahad2013toward, auld2007bayesian}, k-nearest neighbor \cite{Draper-Gil2016, yamansavascilar2017application}, decision trees \cite{Draper-Gil2016, zhang2012network}, bag of words \cite{zhang2012network, zhang2014robust} and cross-entropy similarity \cite{qin2015robust}. Similarly multiple approaches have been studied using different types of neural networks such as: convolutional neural networks \cite{wang2015applications, wang2017end, Lotfollahi2020, Lopez-Martin2017, chen2017seq2img}, recurrent neural networks \cite{Lopez-Martin2017}, probabilistic neural networks \cite{sun2010traffic} and stacked auto-encoders \cite{wang2015applications, Lotfollahi2020}.

Regarding accuracy and precision over encrypted traffic, the state-of-the-art results are the Deep Packet approach~\cite{lotfollahi2018deep} from 2018 (published in~\cite{Lotfollahi2020}) and the Flowpic approach~\cite{Shapira2019} from 2019. Both build up over the works of Draper-Gil et al.~\cite{Draper-Gil2016}, and Lashkari et al.~\cite{lashkari2017characterization} which set up datasets for traffic categorization, the former containing both VPN and unencrypted traffic, and the latter containing both Tor and unencrypted traffic.

In~\cite{Lotfollahi2020} each packet (header and payload) is converted to a normalized byte sequence and used as an input for a 1-D convolutional neural network and a set of stacked auto-encoders. Both are trained with unencrypted and VPN-encrypted traffic. Surprisingly, the resulting networks can categorize traffic for both unencrypted and VPN traffic without using time-related features (but size). Performance is quite good with 94\% average recall and 93\% average recall. Unfortunately, for the unencrypted vs. Tor dataset, there is an important drop in average recall (down to 57\%) and average precision (down to 44\%).

The Flowpic approach~\cite{Shapira2019} uses all spatial and temporal information contained in a traffic flow without looking at its content. Flow packets, encrypted or not, are represented with a 2D-histogram, in the form of a square picture, with packet size in ordinate and normalized arrival times in abscissa. The main idea of Flowpic is that with this approach, the traffic category can be derived from image recognition techniques. The images are thus fed to a LeNet-5 \cite{lecun1998gradient} convolutional neural network, and categorization is done in a class vs. all game with 97\% average accuracy for unencrypted traffic, 99.7\% for VPN traffic, and 85.7\% for Tor traffic. Given their nice results over all three sorts of traffic, Flowpic will be our main point of comparison.

% ===================================================

\section{Proposed model}
\label{sec:Proposed model}

\subsection{Prospects}
\label{sec:Prospects}
The approach used in this paper is very similar to the one employed in \cite{Shapira2019}. Histograms represent flow packets grouped by size and feed the neural network. The difference relies on using an SNN to operate such classification instead of a CNN. No previous research with that method has been reported, and the results will be mainly compared to the study presented in \cite{Shapira2019}. To ensure that this comparison is fair, we follow the same dataset generation and processing as in \cite{Shapira2019}.

\subsection{Dataset generation}
\label{sec:Dataset generation}
To generate the data, we used two public datasets from the University of New Brunswick (UNB) called ``2016 ISCX VPN-nonVPN traffic dataset'' from~\cite{Draper-Gil2016} and ``2017 ISCX Tor-nonTor dataset'' from~\cite{lashkari2017characterization}, both used in the closely related works~\cite{Lotfollahi2020, Shapira2019}. They comprise a range of packet captures (pcap) format files corresponding to traffic flow captures of five main classes (Video, VoIP, Chat, File Transfer, and Browsing) using three different types of encryption (Unencrypted, VPN, and Tor). We converted them into comma-separated values (CSV) files containing arrival times and sizes to exploit the data. We have not considered any restriction about the number of packets or their size. Again it is important to note that there is no Browsing-VPN data in the ISCX dataset, so we have not used this type of data for training or testing our model (as neither did the authors of~\cite{Draper-Gil2016,lashkari2017characterization, Lotfollahi2020, Shapira2019}). Introducing such traffic and making a fair comparison with other approaches when it is included is left as future work.

\subsection{Data processing methodology}
\label{sec:Data processing methodology}
\begin{figure}[]
\centering
\includegraphics[%
width=1.1\textwidth]{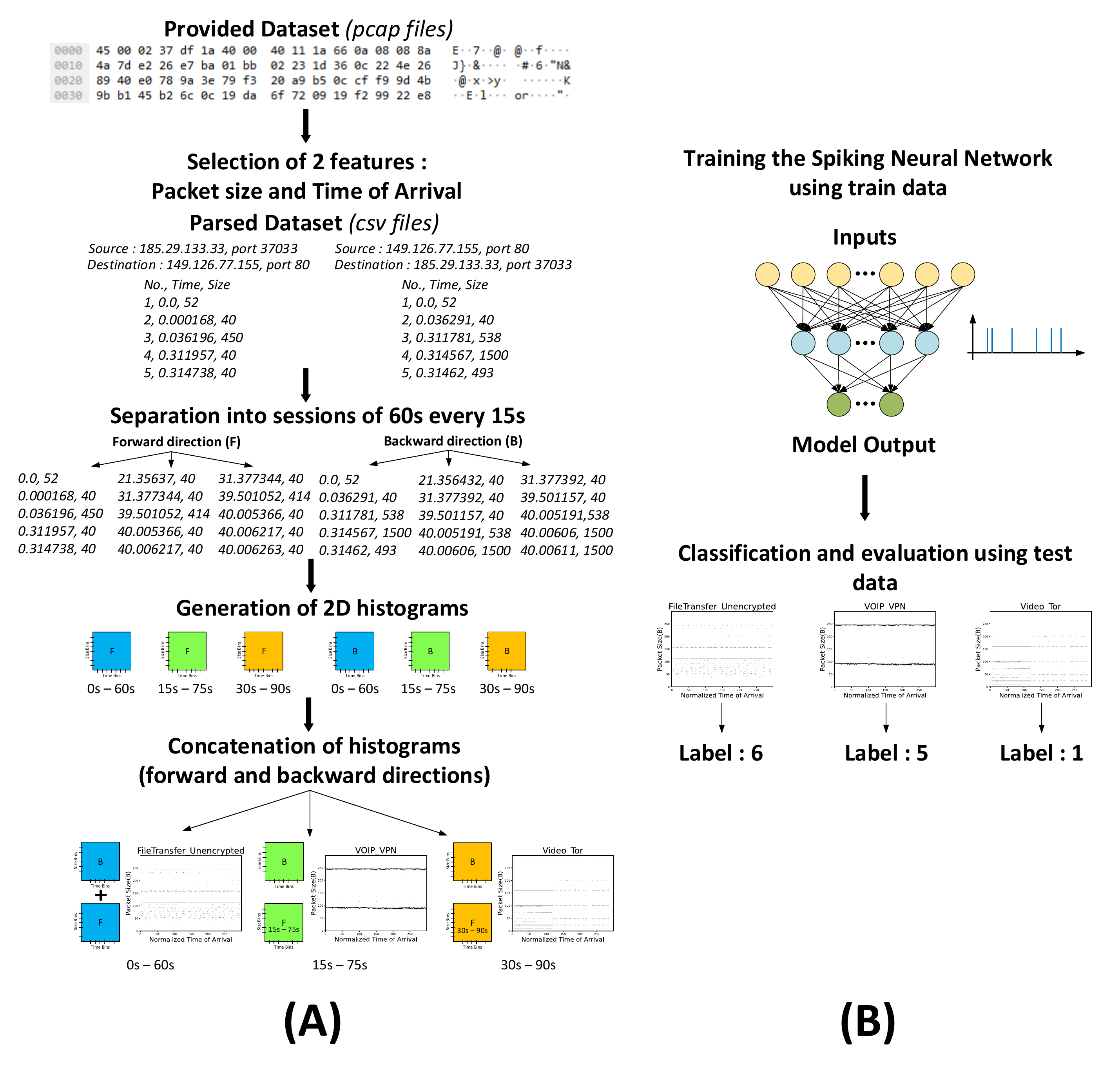}
\caption{\label{fig:work_process_1}Architecture proposed for classifying Internet traffic using a SNN.
Section (A). This section is the part that is similar to \cite{Shapira2019}; at first, from the original pcap files in the dataset, we generate csv files containing the time of arrival and the size of each packet in each flow (forward and backward direction). We separate these sessions into sessions of 60s with 15s of gap between them. Next, we generate two-dimensional histograms using these sessions. The time bins are considered on the X-axis and the size bin on the Y-axis. We also concatenate the forward and backward traffic histograms to generate the final histogram as the input of our network.
Section (B). This section describes our model. The concatenated histograms are fed to the SNN, and the network is trained using these histograms. Finally, the evaluation of the model is done using test dataset histograms.}
\end{figure}

\begin{figure}[]
\centering
\includegraphics[%
width=1.0\textwidth]{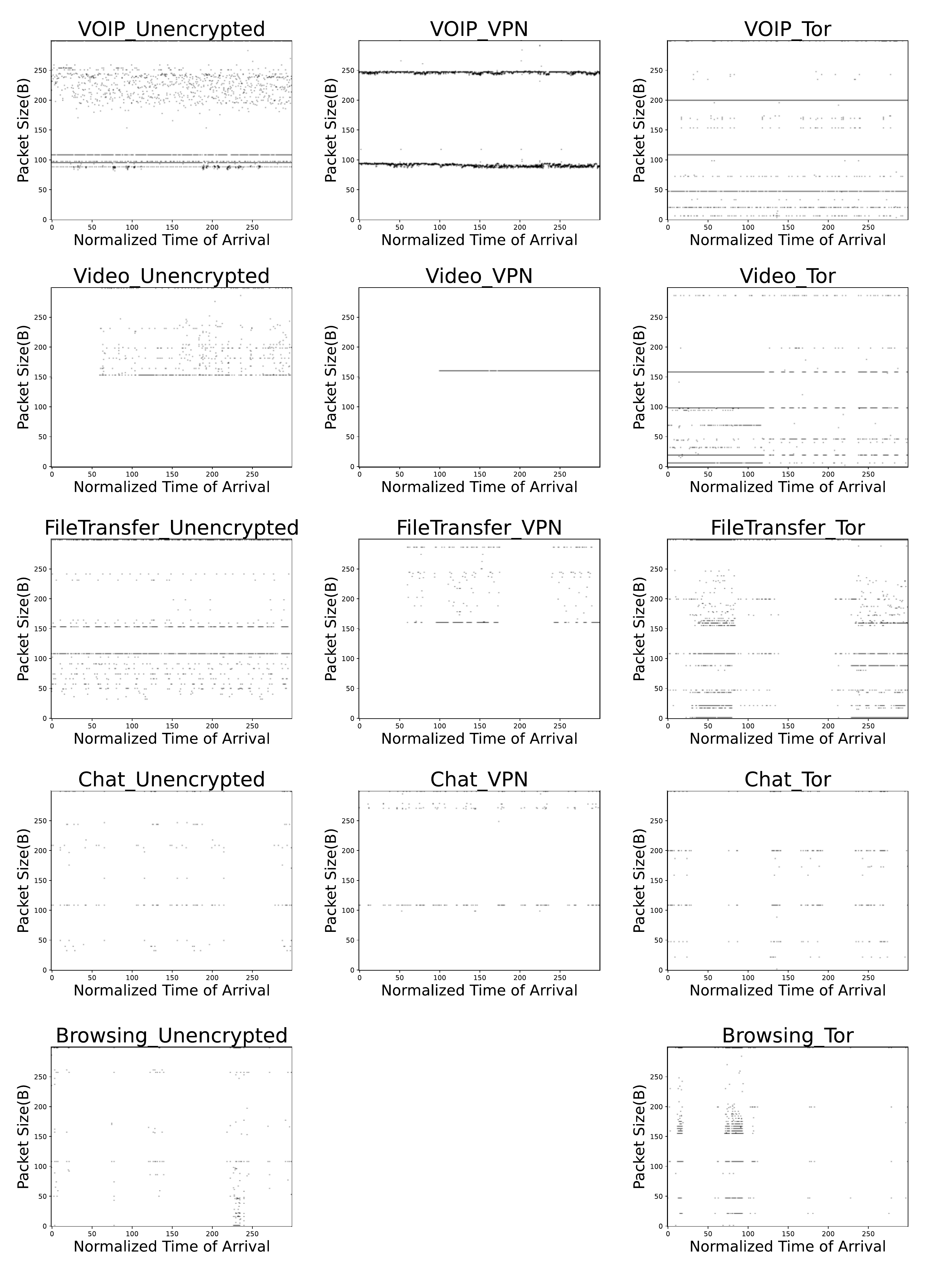}
\caption{\label{fig:Dataset_hist_example_Vertical}Examples of obtained two dimensional histograms for each traffic category. The X-axis and Y-axis show time and size bins, respectively. For clarity, all non-empty bins are represented in black, whatever the packet count. Note: there is no Browsing-VPN data in the ISCX dataset.}
\end{figure}

The first step is to extract the suitable features from these flow packets. A traffic flow can be seen as a sequence of packets with the same source and destination IP. They are bidirectional: forward when a message is sent from source to destination, and backward when a message or acknowledgment (ACK) is sent back from destination to source. Thus, for each exchange, we consider data from both directions of the flow. We pick, as in \cite{Shapira2019}, packet size and time of arrival as features, since they are always available, whether the traffic is encrypted or not. Then, we split each flow every 15s into equal sequences lasting 60s. \revision{It would be helpful to increase even the number of samples of the dataset by splitting the flows into smaller sequences, but there are silent periods of a few seconds in the data, so it is impossible to split the data with much less than 15s steps as it leads to samples without any helpful information.} We have, therefore, csv files representing one-minute chunks of a communication session and containing packet sizes and times of arrival. As a result, we generate 2D-histograms of packet sizes as a function of the time of arrival from each csv file as shown in Figure \ref{fig:work_process_1}. We have two series of histograms for one flow, each corresponding to one direction (forward and backward), concatenated along the y (size) dimension. The histogram values correspond to the number of packets with the same (time size) coordinates. In addition, the resolution of the histograms can be adjusted by increasing or decreasing the number of bins it contains. If the number of bins is too high, the memory and computation requirements for the classification training become prohibitive.
Moreover, as we will associate one input line to each size bin, overfitting can become an issue. Indeed, adding one size bin increases the number of trainable weights by the number of hidden neurons, here 100. If the number of bins is too low, it degrades the histogram quality and, therefore, the network’s ability to recognize the right traffic class. By trial and error, we found that 300 time bins and 300 size bins are a good choice (whereas \cite{Shapira2019} used 1,500x1,500).
Once implemented, the histograms can be classified according to their category. Some examples of images obtained are shown in Figure \ref{fig:Dataset_hist_example_Vertical}.

\subsection{Labeling dataset}
\label{sec:Labeling dataset}
A label is attributed to each csv file according to its traffic category. A label is a number giving both the traffic class (Chat, Browsing, File Transfer, Video or VoIP) and the encryption method (Tor, VPN, or Unencrypted). For instance, label two matches with Video over VPN traffic while label seven matches with File Transfer over Tor traffic. Once created, the csv files containing the histograms are distributed between groups of equal size, called batches, used as input data for the neural network training. Thus, the batch size, representing how many histograms each batch contains, is a parameter to be adjusted. Increasing it avoids scattering data, whereas decreasing it introduces stochasticity, benefiting the gradient descent.
Moreover, we use a weighted random sampler on our training dataset. For each class, a weight is attributed to each file, which is inversely proportional to the number of files in this class. This allows to balance the batches in the training dataset and therefore to improve the classification performance.\\

% \begin{table}[h]
% \centering
% \begin{tabular}{|p{2cm}|p{2.5cm}|p{2.5cm}|p{2.5cm}|} \hline
%  & unencrypted & Tor & VPN \\\hline
% Browsing & Firefox and Chrome & Firefox and Chrome & None \\\hline
% Chat & AIM Chat, Facebook, Google Hangouts, ICQ, Skype, WhatsApp Web & AIM Chat, Facebook, Google Hangouts, ICQ, Skype & AIM Chat, Facebook, Google Hangouts, ICQ, Skype \\\hline
% File Transfer & FTP, POP, SCP, SFTP, Skype & FTP, Skype, SFTP & FTPS, Skype, SFTP \\\hline
% Video & Facebook, Google Hangouts, Netflix, Vimeo, YouTube & Vimeo, YouTube & Netflix, Vimeo, YouTube \\\hline
% VoIP & Facebook, Google Hangouts, Skype, VoIP Buster & Facebook, Google Hangouts, Skype & Google Hangouts, Skype, VoIP Buster\\\hline
% \end{tabular}
% \caption{\label{tab:Dataset}Dataset\cite{Shapira2019}}
% \end{table}

\subsection{The model}
\label{sec:The model}
\subsubsection{Spiking Neural Network}
\label{sec:Spiking Neural Network}
\begin{figure}[]
\centering
\includegraphics[%
width=0.7\textwidth]{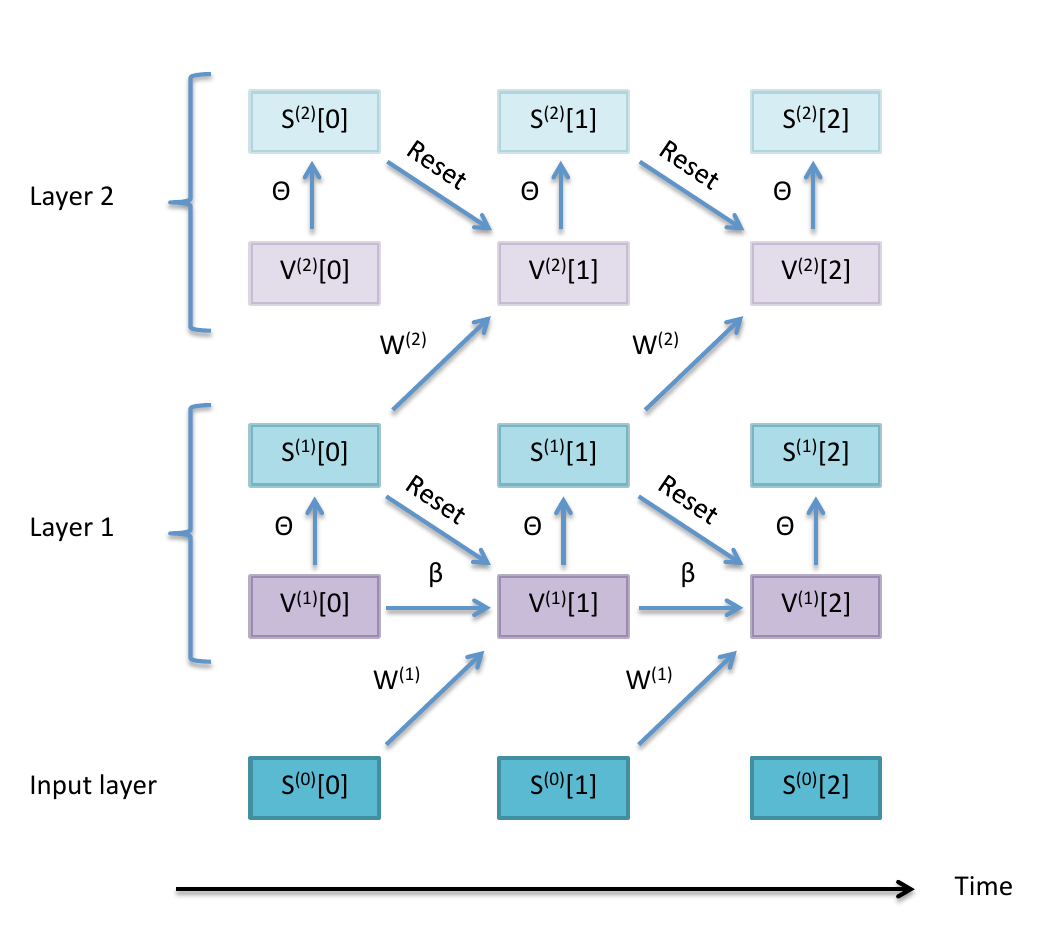}
\caption{\label{fig:BPTT}\revision{The unfolded Spiking Neural Network}. This image shows the unfolded SNN demonstrating how the gradient is backpropagated through the network using BPTT.\cite{Neftci2019}}
\end{figure}

\begin{figure}[]
\centering
\includegraphics[%
width=1.0\textwidth]{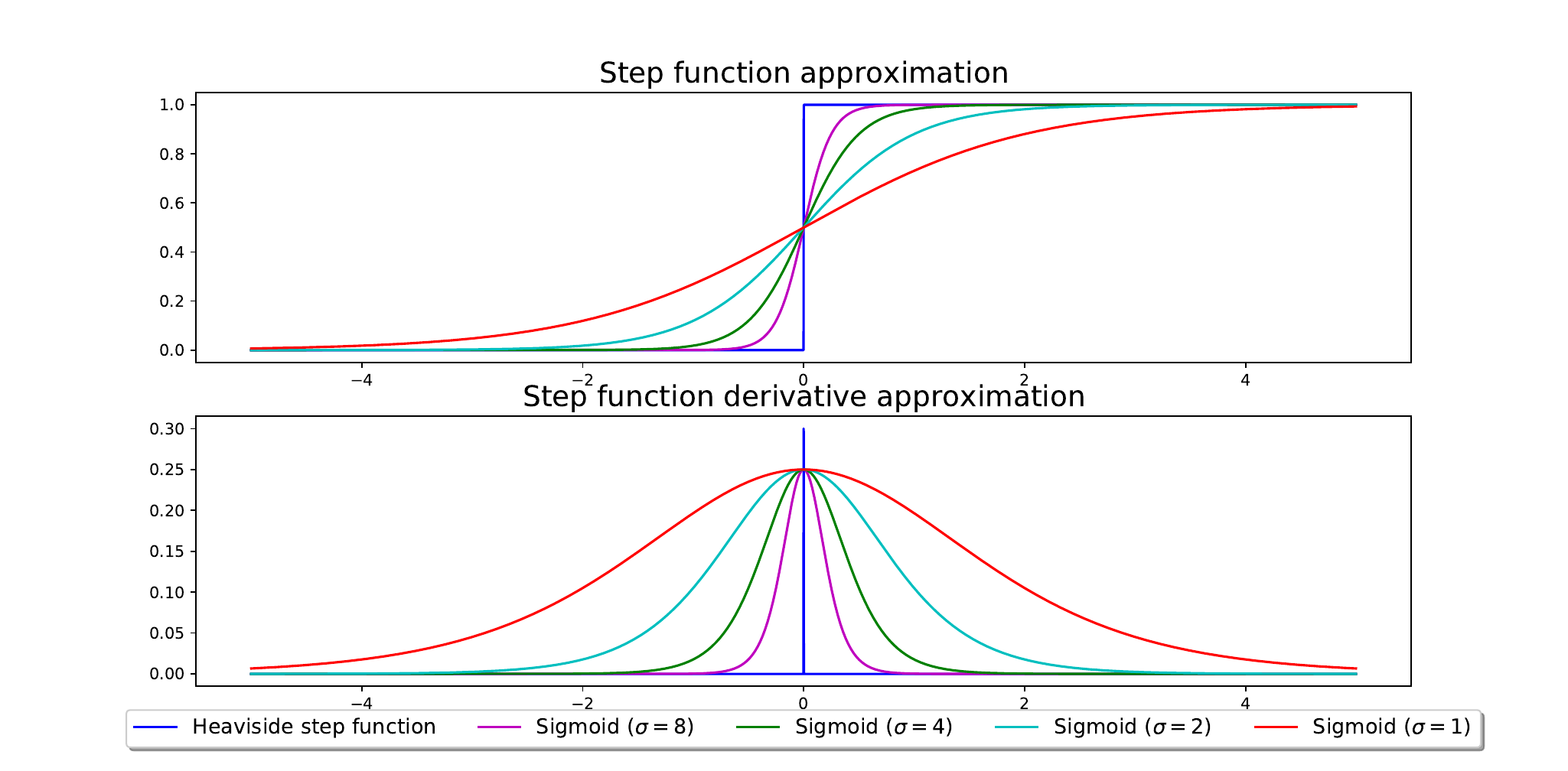}
\caption{\label{fig:SG_heaviside}Surrogate gradient for activation function. Different Sigmoid functions can be a suitable approximation for the Heaviside function. Because the step function is not differentiable and has no gradient, we need to use a surrogate gradient in the backpropagation algorithm of training the SNN, so the sigmoid function is assumed as a substitution of the main Heaviside function. Its gradient is used as a surrogate gradient in the backpropagation algorithm. In the forward path, we use the main Heaviside function, but the surrogate gradient is used in the backward path.\cite{Neftci2019}}
\end{figure}

An SNN can be seen as a type of biologically inspired deep learning model whose state evolves in time and in which spikes transmit information between neurons \cite{Neftci2019}. In most SNNs, neurons are modeled using the leaky integrate-and-fire (LIF) model. It introduces a neuron dynamic able to capture temporal patterns. In discrete-time, it can be described as follows:
\begin{equation}
V[n+1]= \beta (V[n]-\revision{R[n]}) + (1-\beta) \sum_j w_j S_j^{in}[n]
\label{eq:1}
\end{equation}
\begin{equation}
S^{out}[n+1]= \theta(V[n+1]-\revision{V_{th}})
\label{eq:2}
\end{equation}
\begin{equation}
R[n+1]=V_{th}.S^{out}[n]
\label{eq:3}
\end{equation}
$V$ represents the membrane potential, n the discrete time, $\beta=e^{-\frac{\Delta t}{\tau}}$ the leak coefficient (with $\tau$ the membrane time constant),$\sum_{j}w_jS_j^{in}$ the weighted \revision{sum of} input spikes, $R$ the reset function due to the output spikes, $S^{out}$ the output spike, \revision{$V_{th}$ the neuron's threshold voltage}, and $\theta$ the Heaviside step function. Equation \ref{eq:2} implies that whenever the membrane potential reaches the threshold, an output spike is fired and the potential is reset.
As for ANNs, an SNN can be described as a large number of neurons connected via weighted synapses that are represented by weights. A weight is a number that evaluates how information, here spikes, is transmitted between two neurons. Thus, training a network means adjusting the weights so that data provided as input leads to the activation of the correct output. The difference between the result provided by the network and the desired result is computed by a loss function and has to be minimized. Here, we use the cross-entropy function $H$ calculated as follows
\begin{equation}
H(p,q)= -\sum_i p_ilog(q_i)
\label{eq:4}
\end{equation}
With $p_i=1$ if the true label is $i$ and 0 otherwise, and $q_i$ the predicted value of the model.
To adjust the weights and train the network, the mainly used method that has proved to be efficient is backpropagation. It consists of updating the weights from the output layer back to the network's input layer in a way that minimizes the loss function. As SNN are recurrent (see Eq. 2), backpropagation through time (BPTT), which consists in unfolding through time the network before applying backpropagation, is employed (see Figure \ref{fig:BPTT}).
Still, this method requires to calculate the gradient of the activation function. SNNs pose an issue because the Heaviside function gradient is 0 everywhere except in 0, which implies that no information can flow through the activation function during the backpropagation. Recent research has managed to solve this problem by switching the true gradient by a surrogate function \cite{Neftci2019}. Here we used the derivative of a sigmoid since a sigmoid is continuous and can approximate the Heaviside function as seen in Figure \ref{fig:SG_heaviside}.

\begin{figure}[]
\centering
\includegraphics[%
width=1.0\textwidth]{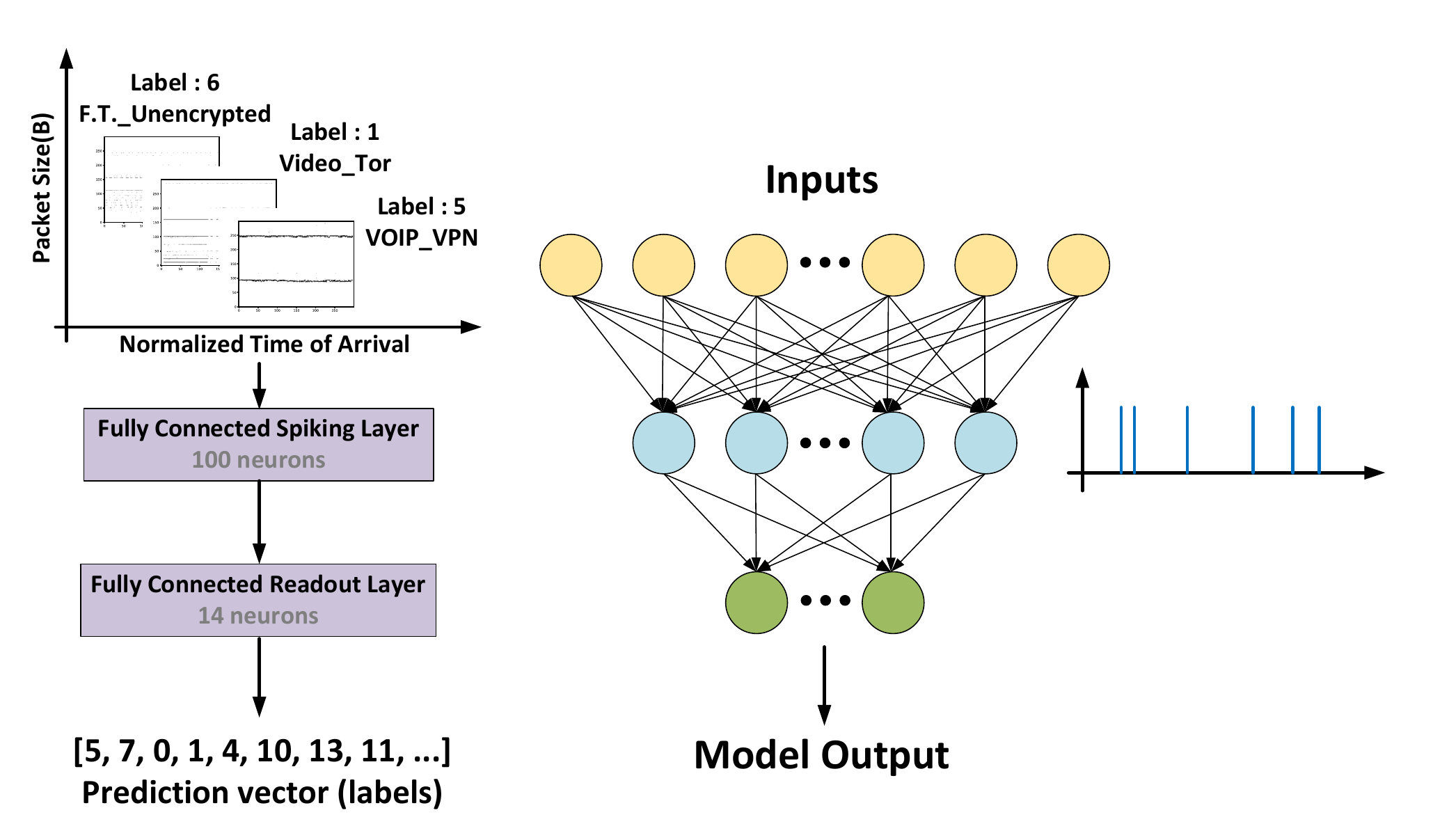}
\caption{\label{fig:Proposed_architecture}Architecture of the proposed network. In the proposed architecture, there is just one hidden layer with 100 neurons and one Readout layer with 14 neurons. The size of input histograms fed to the network (in the input layer) is (300,300) which is 300 for time bins and 300 for size bins on the 2D-histogram. The time step of the spiking neurons is set to the time bins dimension, and the input layer size is set to the size bin dimension. At the Readout layer (with 14 neurons), each neuron is representative of one traffic category, and the winner neuron is the neuron with the highest mean potential during all time steps. The winner neuron tells which class is chosen by the network.}
\end{figure}

One of the main assets of SNNs compared to other networks is that inference can be implemented at low power using event-driven computation on neuromorphic chips \cite{Roy2019}. They are thus appealing for embedded systems such as satellites. Furthermore, SNNs are adapted to dynamic inputs: they detect synchronicity and other time-related features.
The proposed architecture for traffic classification consists of two fully connected layers: one Spiking Dense Layer made up of 100 neurons with 300 inputs and one Read Out Layer with 14 neurons corresponding to the different categories to classify (Figure \ref{fig:Proposed_architecture}). We adapted the s2net code (available at \url{https://github.com/romainzimmer/s2net}), designed to train both fully connected and convolutional SNNs using surrogate gradient learning, and based on PyTorch. This code has already given excellent results on the Google speech commands dataset \cite{Zimmer2019, Pellegrini2021}, and for epileptic crisis detection from EEG signal \cite{Soltani2020}. In our case, the network receives histogram batches as inputs and then delivers a prediction vector over the 14 labels (corresponding to the 14 categories in Figure \ref{fig:Dataset_hist_example_Vertical}). The first layer is a dense spiking layer. The neurons' potentials (V) are calculated for the 300 time steps, and spikes are generated according to equation \ref{eq:2}. The second dense layer (called the Readout layer) can be seen as a spiking layer with an infinite threshold. The potential of each neuron is calculated over the 300 times steps using equation \ref{eq:2}, and then the neuron with the highest mean potential is the winner and specifies the prediction of the network for the corresponding input.

% ===================================================

\section{Experiments}
\label{sec:Experiments}

\revision{Our dataset has a total of 17,450 data samples before being divided into different sets.} The dataset has been then split into three separated sets: 65\% of samples used to train the model, 15\% to validate it during the training phase, and 20\% to evaluate it during the testing phase. \revision{Before splitting, the number of data samples in different categories is a few hundred for Video/File transfer/Chat in each of Non-VPN/VPN/Tor techniques and a few thousand in the case of VoIP/Browsing.} The training phase consists of optimizing the network by adjusting its weights, but also the leak coefficients $\beta$ (we had one trainable $\beta$ per layer, as previous works suggested that training these coefficients can increase accuracy \cite{Fang2020b, Yin2020}). \revision{Also, the value of neuron's threshold $V_{th}$ per layer is learned by the gradient descent.} The validation set is used to control model performances and tuning hyper-parameters, whereas the testing phase enables the evaluation of final model performance. As a result, we extracted independent data from the dataset used only for the test. Otherwise, there could be a risk of developing an over-fitted model. This would be an important issue since it would mean the model is biased, too specific, and unable to classify new data correctly.
For training, we set the batch size to 128 and used 30 epochs. Cross entropy loss was used as the loss function. We also used the \emph{Adam} optimizer along with \emph{ReduceLROnPlateau} as a scheduler which allows us to adjust the learning rate according to the epoch and thus to improve gradient descent for BPTT (the initial value of learning rate was 5e-4). To avoid over-fitting, we also introduce a regulatory loss coefficient that penalizes solutions using a large number of spikes. Thus, even if classification may be less efficient, the number of spikes remains quite low, saving energy. This could be very significant in embedded systems, for instance.
In the experiments, we have tested several values for different hyper-parameters of our model, which are explained in Table \ref{tab:hyper_parameters_range}. \revision{The mentioned hyper-parameters are tuned by exhaustive search.}
We have used Nvidia Tesla T4 GPU for final experiments, and the average run time is 131 seconds for each epoch of training and 19 seconds for evaluation on the test dataset.

% \begin{table}[]
% \centering
% % \begin{center}
% \begin{tabular}{ |m{3cm}|m{1cm}|m{1cm}|m{1cm}| } 
%     \hline
%     \textbf{Type/Encryption} & \textbf{Non-VPN} & \textbf{VPN} & \textbf{Tor} \\\hline
%     \textbf{VoIP} & 3590 & 1433 & 1489 \\\hline
%     \textbf{Video} & 986 & 302 & 694 \\\hline
%     \textbf{File Transfer} & 729 & 242 & 563 \\\hline
%     \textbf{Chat} & 333 & 620 & 214 \\\hline
%     \textbf{Browsing} & 5249 & - & 1006 \\
%     \hline
% \end{tabular}
% % \end{center}
% \caption{\label{tab:dataset_samples_number}\revision{Total number of data samples in different traffic types and encryption techniques. These samples then are divided to Training, Validation and Test data with the mentioned proportions.}}
% \end{table}

\begin{table}[]
\centering
% \begin{center}
\begin{tabular}{ |m{4cm}|m{4cm}| } 
    \hline
    \textbf{Hyper-parameter} & \textbf{Range of tested values} \\\hline
    Number of dense layer neurons & 20, 40, 60, 80, \textbf{100}, 120, 140, 160, 200, 250, 300 \\\hline
    Number of time bins in histograms and number of time steps of neurons & 100, 200, \textbf{300}, 400, 500, 750, 1,000, 1,500, 2,000, 2,500\\\hline
    Number of packet size bins in histograms & 100, 200, \textbf{300}, 400, 600 \\
    \hline
\end{tabular}
% \end{center}
\caption{\label{tab:hyper_parameters_range}Range of the hyper-parameters tested in our experiments. The Bold values show the value with the best final test accuracy. \revision{The best values are found by exhaustive search.}}
\end{table}

% ===================================================

\section{Results}
\label{sec:Results}
\begin{figure}[]
\centering
\includegraphics[%
width=1.0\textwidth]{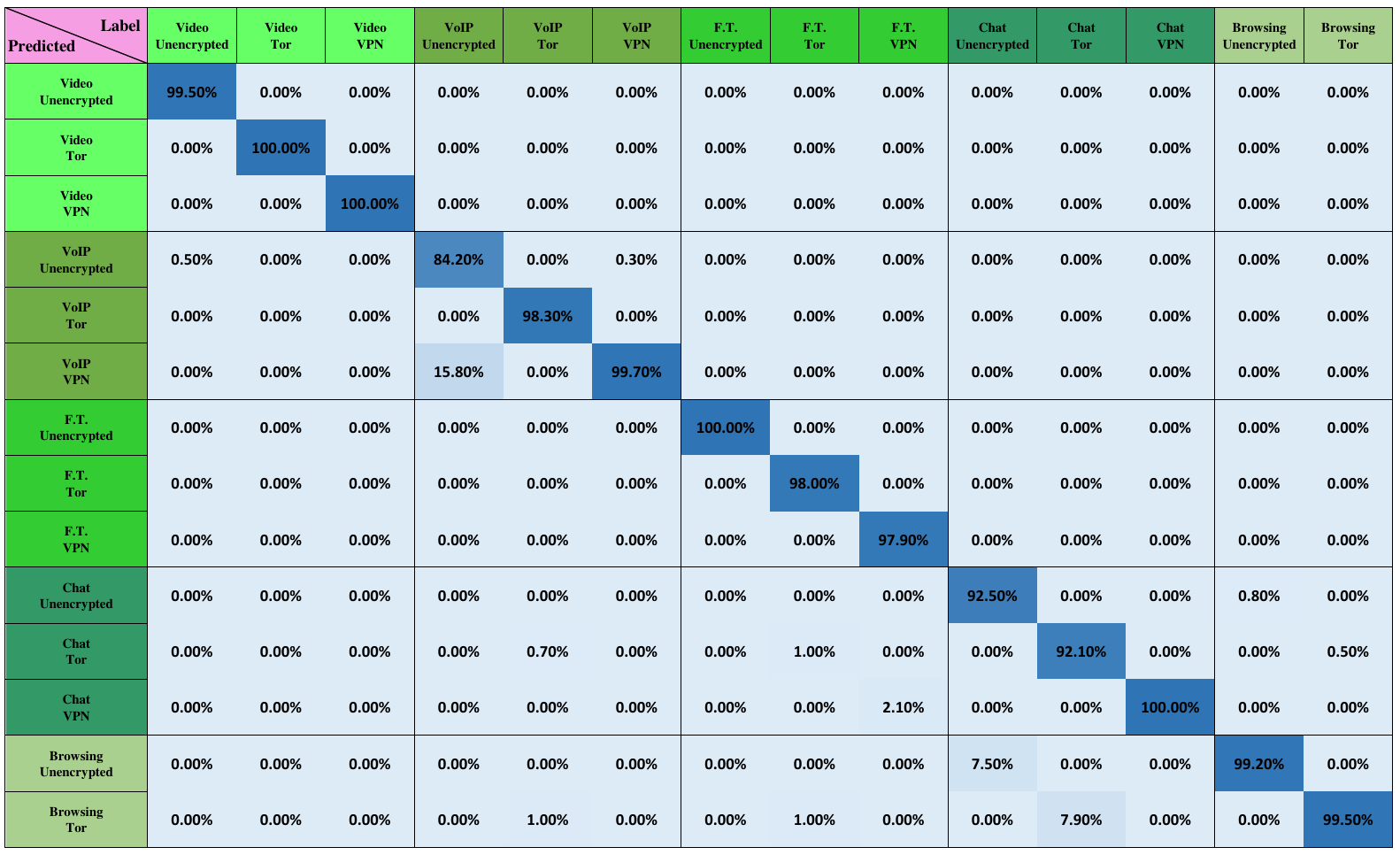}
\caption{\label{fig:confusion_matrix}Confusion matrix of the proposed model on the categorization of all classes simultaneously (including all the data from different traffic classes and encryption methods and 14 classes as depicted in the matrix). The classes on the lines are the predicted labels while the classes on the columns are the accurate labels}
\end{figure}

As explained in Section \ref{sec:Labeling dataset} our model predicts both the traffic category (e.g., VoIP) and the encryption technique (e.g., VPN) at the same time, and it was trained over the whole dataset. This is in contrast with \cite{Shapira2019}, where the dataset was split in three according to the encryption techniques, and different classifiers were trained for each encryption technique and for each task (``one vs. all'' or multi-class categorization). The task in \cite{Shapira2019} is thus simpler as each of their three classifiers only needs to find one among five categories (or four in the VPN case), whereas in our case, it must find one among 14. That being said, we can compute the accuracy of our network for all the tasks used in \cite{Shapira2019}, because they are sub-tasks of the tasks we solved (i.e., all performance measures can be computed from the confusion matrix of Figure \ref{fig:confusion_matrix}). However, it is important to note that not only does the proposed model outperform Flowpic at its sub-tasks, but it also solves a more complicated problem with a very small error rate.

We first evaluated our model in one vs. all classification, meaning that each class is compared to the rest of the dataset. We have used three different performance metrics. For one class, recall (Re) measures the model's capacity to successfully retrieve the class and not to miss it. Precision (Pr) measures the ability of the model to specifically retrieve the class and ensure the class returned is the good one. Accuracy (Ac) represents a global score quantifying if the model is often right or not.
\begin{equation}
Pr = \frac{TP}{TP+FP}
\label{eq:5}
\end{equation}

\begin{equation}
Re = \frac{TP}{TP+FN}
\label{eq:6}
\end{equation}

\begin{equation}
Ac = \frac{TP+TN}{TP+TN+FP+FN}
\label{eq:7}
\end{equation}

\begin{figure}[]
\centering
\hspace{-3em}~\includegraphics[%
width=1.0\textwidth]{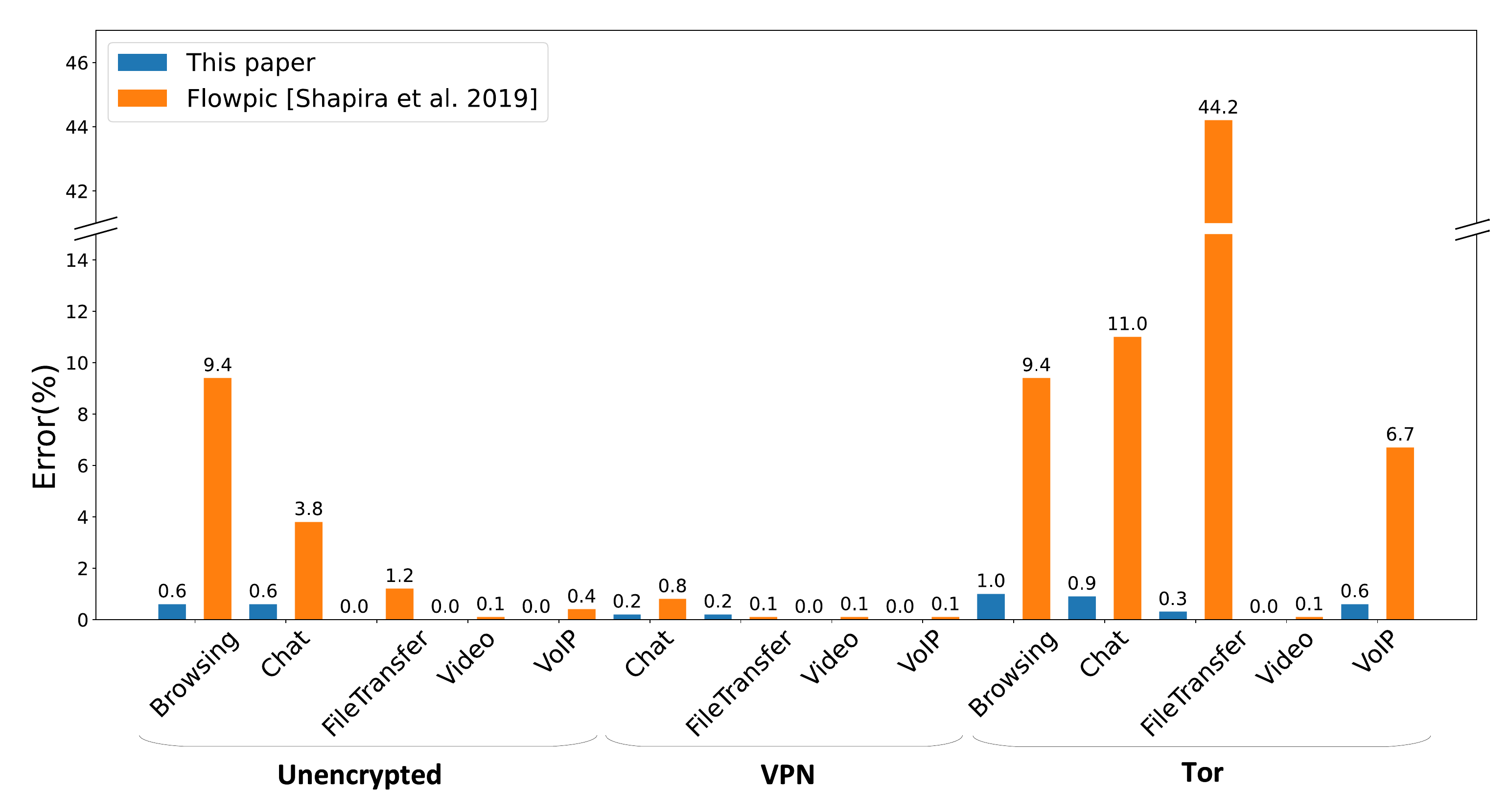}
\caption{\label{fig:Error_comparison_total}Comparison of one vs all errors on categorization of different traffic classes and encryption methods between this work and Flowpic \cite{Shapira2019}. Note that in Flowpic, the unencrypted traffic is named Non-VPN. The task was one versus all classification of different traffic categories with the data from just one encryption method (Unencrypted, VPN, Tor). We have used the error rate on classification (100-accuracy) of the data instead of the accuracy to highlight the relative multiplicative gap between the performance of the proposed network in comparison to Flowpic \cite{Shapira2019}. As the Figure depicts, this paper is better in all traffic classes and encryption methods except File transfer-VPN which Flowpic performs slightly better.}
\end{figure}

\begin{figure}[]
\centering
\hspace{-2em}~\includegraphics[%
width=1.0\textwidth]{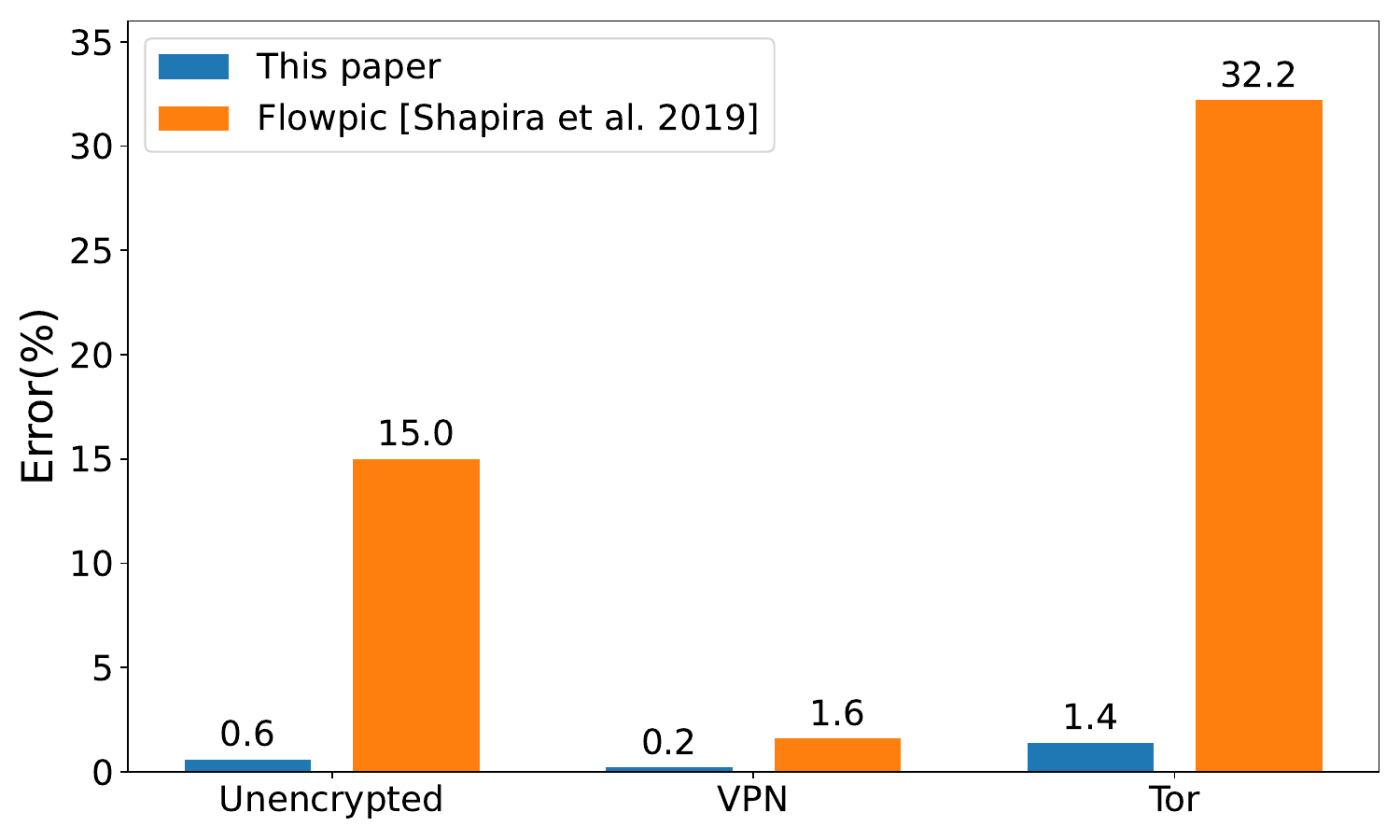}

% \pgfplotstableread[row sep=\\,col sep=&]{
%     Encryption_type     & Ours & Flowpic \\
%     Unencrypted              & 96.4 & 85 \\
%     VPN                 & 96.4 & 98.4 \\
%     TOR                 & 99.9 & 67.8 \\
%     }\EncryptionData

% \begin{tikzpicture}
% \centering
%     \begin{axis}[
%             ybar,
%             bar width=.5cm,
%             width=\textwidth,
%             height=.5\textwidth,
%             legend style={at={(0.5,1)},
%                 anchor=north,legend columns=-1},
%             symbolic x coords={Unencrypted,VPN,TOR},
%             xtick=data,
%             nodes near coords,
%             nodes near coords align={vertical},
%             ymin=60,ymax=115,
%             ylabel={Accuracy(\%)},
%         ]
%         \addplot table[x=Encryption_type,y=Ours]{\EncryptionData};
%         \addplot table[x=Encryption_type,y=Flowpic]{\EncryptionData};
%         \legend{Ours, Flowpic}
%     \end{axis}
% \end{tikzpicture}
\caption{\label{fig:Error_comparison_traffic_categorization_on_encryption}Comparison of error rates on categorization of different encryption techniques between this work and Flowpic \cite{Shapira2019}. The task was multi-class categorization between different traffic classes of each encryption method. We outperformed Flowpic for all encryption methods. }
\end{figure}

% \\\\
% $Pr = \frac{TP}{TP+FP}$
% $Re = \frac{TP}{TP+FN}$
% $Ac = \frac{TP+TN}{TP+TN+FP+FN}$\\\\
TP, FP, TN, FN stand for True Positive, False Positive, True Negative, and False Negative, respectively.
For the sake of brevity, in this section, we report the accuracy measure, but the other measures can be found in \ref{appendix:Complete results}.
Table \ref{tab:perf_comparison_total} shows the results for each category and encryption technique. It extends Table 4 in~\cite{Shapira2019} with our results (we kept the same form to simplify the comparison). Note, however, that in~\cite{Shapira2019} they also provide accuracy when using a classifier (e.g., VPN-trained) on another setting (e.g., Tor). As we only have one classifier for all the categories, we only kept for the comparison the best figures of the original table (which correspond to using the right classifier in its own setting, e.g., the VPN-trained classifier to classify VPN traffic). We have reached at least 99.0\% of accuracy in all categories and even 100\% for some categories. Moreover, our model outperforms \cite{Shapira2019}'s model for all categories except File transfer-VPN, in which Flowpic slightly performs better (99.9\% vs. 99.8\% for us). Figures \ref{fig:Error_comparison_total} and \ref{fig:Error_comparison_traffic_categorization_on_encryption} compare again the results of our network with \cite{Shapira2019} in categorization tasks on different traffic classes and different encryption techniques using error rates instead of success rates. As the accuracy rates are all very close to $100\%$, it is hard to compare the relative improvements. These graphs are probably a better indicator of the relative multiplicative drop obtained on the number of errors when using our strategy. The proposed model also reached an accuracy of 96.7\% on the multi-class categorization of different encryption techniques, which is better than the 88.4\% reported by \cite{Shapira2019} (see Table 3, row 7).

Finally, we reached an overall accuracy of 95.9\% and average precision of 96.9\% (full precision measures are available in \ref{appendix:Complete results}). Note that \cite{Shapira2019} could not report this global accuracy because they always separated the encryption techniques. Figure \ref{fig:confusion_matrix} shows the confusion matrix. We observe some confusions between, for instance, Chat Tor and Browsing Tor or between VoIP Unencrypted or VoIP VPN. When analyzing the failed instances, we observed that most of them contained too few packets to be classified correctly.

\begin{table}[]
\centering
\begin{tabular}{|c||c|c c c|} \hline

\textbf{Class} & \multicolumn{4}{c|}{\textbf{Accuracy (\%)}} \\\hline \hline
% Class & Accuracy (\%) & & &  \\\hline

\multirow{3}{*}{\textbf{VoIP}} & Work/Data     & Unencrypted   & VPN   & Tor \\\cline{2-5}
 & Flowpic       & 99.6     & 99.9      & 93.3 \\
 & This paper    & 100.0    & 100.0     & 99.4 \\\hline \hline

\multirow{3}{*}{\textbf{Video}} & Work/Data  & Unencrypted   & VPN   & Tor \\\cline{2-5}
 & Flowpic       & 99.9     & 99.9      & 99.9 \\
 & This paper    & 100.0    & 100.0      & 100.0 \\\hline \hline

\multirow{3}{*}{\textbf{File Transfer}} & Work/Data  & Unencrypted   & VPN   & Tor \\\cline{2-5}
 & Flowpic       & 98.8     & 99.9      & 55.8 \\
 & This paper    & 100.0    & 99.8      & 99.7 \\\hline \hline

\multirow{3}{*}{\textbf{Chat}} & Work/Data  & Unencrypted   & VPN   & Tor \\\cline{2-5}
 & Flowpic       & 96.2     & 99.2      & 89.0 \\
 & This paper    & 99.4     & 99.8      & 99.1 \\\hline \hline

\multirow{3}{*}{\textbf{Browsing}} & Work/Data  & Unencrypted   & VPN   & Tor \\\cline{2-5}
 & Flowpic       & 90.6     & -         & 90.6 \\
 & This paper    & 99.4     & -         & 99.0 \\\hline
 
\end{tabular}
\caption{\label{tab:perf_comparison_total}Comparison between results of this paper and Flowpic \cite{Shapira2019} in one vs all categorization of different traffic classes. It is noteworthy that the unencrypted traffic is named Non-VPN in Flowpic~\cite{Shapira2019}. As the table depicts, the results are improved in all traffic classes and encryption methods except File transfer-VPN, in which Flowpic performs slightly better.}
\end{table}

% \begin{figure}[H]
% %\centering
% \hspace{-2em}~\includegraphics[%
% width=1\textwidth]{Error_comparison_encryption_multiclass}
% \caption{\label{fig:Error_comparison_encryption_multiclass}Comparison of error rates on multi-class categorization of different encryption techniques between this work and Flowpic \cite{Shapira2019}. We outperformed FlowPic on this task.}
% \end{figure}

% \begin{table}[h]
% \centering
% \begin{tabular}{|p{3.5cm}|p{2.5cm}|} \hline
% \textbf{Traffic classes} &	\textbf{Accuracy(\%)} \\\hline
% Browsing-unencrypted &       97.4 \\\hline
% Browsing-Tor &      	99.4 \\\hline
% Chat–unencrypted &       	99.2 \\\hline
% Chat–Tor &          	99.5 \\\hline
% Chat–VPN &          	99.9 \\\hline
% File Transfer–unencrypted &	98.9 \\\hline
% File Transfer–Tor & 	99.9 \\\hline
% File Transfer–VPN & 	99.9 \\\hline
% Video–unencrypted &      	99.4 \\\hline
% Video–Tor &         	99.8 \\\hline
% Video–VPN &         	99.9 \\\hline
% VoIP-unencrypted &       	94.8 \\\hline
% VoIP–Tor &          	99.9 \\\hline
% VoIP-VPN &          	95.9 \\\hline
% \end{tabular}
% \caption{\label{tab:total_perf}list of performances in different traffic classes and encryption techniques}
% \end{table}

The results highlight that our SNN can classify different traffic classes with very good performance according to three encryption techniques. This is even more noticeable considering the limited amount of data needed for training and the simplicity of our network. It is fully connected with only one hidden layer compared to the two convolutional layers plus one fully connected layer used in \cite{Shapira2019}. This can be very useful for embedded systems such as satellites due to their low energy requirements.

Finally, we performed some additional analyses to get more insight into why our SNN reaches such a good accuracy despite its simplicity. Firstly, we tried to destroy all the temporal features at inference time (after regular training). This was done by shuffling the histograms along the temporal dimension. More specifically, we randomly permuted the elements of each row of the histograms, using a different permutation for each row. This completely destroys the temporal features, but not the spatial ones (for example, if a specific sort of traffic tends to use larger packets than another, it can still be used for discrimination). Such an approach led to a total accuracy of only 90.6\%, and the comparison to the accuracy in normal mode (95.9\%) indicates that our SNN does exploit temporal features.

Secondly, we tried to destroy all the temporal features but the synchronous ones (still at inference time, after regular training). This was done by reshuffling the histograms along the temporal dimension, but this time the same permutation was used for all the rows of a given histogram. This means that if multiple packets with different sizes arrived in the same time bin before the shuffling, they still arrived in the same time bin after it. Conversely, any inter-time-bins temporal information, i.e., long-range temporal correlations, was lost. Of course, this second shuffling method also conserved the spatial features. This second experiment led to an accuracy of 95.1\%, almost as good as the baseline one (95.9\%). This means that our SNN mainly exploits synchronous features (with a 200 ms resolution), largely ignoring longer-range correlations. Of course, this second sort of shuffling is more harmful when increasing the resolution (e.g., with 40ms time bins, the accuracy drop is about 2\%, indicating that cross-correlations with a 40-200 ms lag matter).

To confirm the range of valuable correlations, we went back to a 200 ms resolution, and we analyzed the final values of the leak coefficients $\beta$ (recall that it is trainable). It turns out that, whatever the initial values, the $\beta$ of the hidden layer converges towards $\sim$ 0.42. With such a fast decay, neurons quickly forget any packet that arrived before the current time bin. In other words, neurons mostly care about synchrony (with a 200ms resolution). In the last experiment, we forced $\beta=0$. In this extreme case, the neurons become stateless: they only care about current inputs. This led to an accuracy of 95.2\%. This is consistent with the second shuffling experiment and confirms that inter-time-bins correlations are not very informative when the time-bin size is large enough to capture short-term synchronicity.

Altogether, these additional experiments confirm that the dataset contains temporal features, mostly synchrony patterns (with a 200ms resolution), that are highly diagnostic and efficiently exploited by an SNN with a fast decay.

% ===================================================

\section{Conclusion}
\label{sec:Conclusion}

In this paper, we proposed an approach to use an SNN to classify different traffic classes encrypted with different techniques (via VPN or Tor). \revision{No previous work has used SNNs for the purpose of traffic classification.} We only considered packet size and time of arrival of flow traffic. Therefore, the input data were 2D-histograms representing a session of 60s in both forward and backward directions. We reached an overall accuracy of 95.93\% and an average precision of 96.9\% on the test dataset with a simple network made up of only one hidden layer and trained on only 65\% of the dataset. Thus, we have shown that SNN can be very relevant for Internet traffic classification with applications in the industry since SNN can run on neuromorphic hardware with low energy consumption \cite{Roy2019}. Finally, progress can be made, and the model performance can be enhanced by considering spiking convolutional layers that are very promising.

% ===================================================

\section{Acknowledgements}
We thank the authors of the Flowpic paper \cite{Shapira2019}, Tal Shapira and Yuval Shavitt, for sharing their processed pcap files with us.

% ===================================================

\appendix
\section{Complete results}
\label{appendix:Complete results}
In this section, we bring complete performance results, while in Section~\ref{sec:Results} we just mentioned the accuracy. Complete results of one vs all categorization are presented in Table \ref{tab:traffic_class_categ_perf_complete} for different traffic classes, and in Table \ref{tab:all_traffic_categ_perf_complete} for all different traffic classes and encryption methods. The accuracy and average precision of the network are respectively 95.93\% and 96.9\% in the multi-class categorization of all types of traffic (traffic classes and encryption methods simultaneously with 14 output neurons). Finally, the average accuracies for one vs. all categorization of different traffic classes with the data of each encryption technique are reported in Table \ref{tab:encryption_categ_perf_complete}.

\begin{table}[H]
\centering
\begin{tabular}{|c c c c|} \hline
\textbf{Traffic class} & \textbf{Recall(\%)} & \textbf{Precision(\%)} & \textbf{Accuracy(\%)} \\\hline
Browsing &          99.3 &  99.0 &	99.4 \\\hline
Chat &      	    96.4 &	94.3 &	99.4 \\\hline
File Transfer &     98.9 &	100.0 &	99.9 \\\hline
Video &     	    99.8 &	100.0 &	100.0 \\\hline
VoIP &      	    99.6 &	99.9 &	99.8 \\\hline
\end{tabular}
\caption{\label{tab:traffic_class_categ_perf_complete}Complete performance results on one vs all categorization of different traffic classes. The task was to classify each traffic class versus all other classes with all the data from the test dataset.}
\end{table}

\begin{table}[H]
\centering
\begin{tabular}{|c c c c|} \hline
\textbf{Traffic class} &	\textbf{Recall(\%)} &	\textbf{Precision(\%)} &	\textbf{Accuracy(\%)} \\\hline
Browsing-Unencrypted &       99.2 &	99.5 &	99.6 \\\hline
Browsing-Tor &      	99.5 &	96.6 &	99.8 \\\hline
Chat–Unencrypted &       	92.5 &	88.6 &	99.6 \\\hline
Chat–Tor &          	92.1 &	89.7 &	99.8 \\\hline
Chat–VPN &          	100.0 &	99.2 &	100.0 \\\hline
File Transfer–Unencrypted &	100.0 &	100.0 &	100.0 \\\hline
File Transfer–Tor & 	98.0 &	100.0 &	99.9 \\\hline
File Transfer–VPN & 	97.9 &	100.0 &	100.0 \\\hline
Video–Unencrypted &      	99.5 &	100.0 &	100.0 \\\hline
Video–Tor &         	100.0 &	100.0 &	100.0 \\\hline
Video–VPN &         	100.0 &	100.0 &	100.0 \\\hline
VoIP-Unencrypted &       	84.2 &	99.7 &	96.6 \\\hline
VoIP–Tor &          	98.3 &	100.0 &	99.9 \\\hline
VoIP-VPN &          	99.7 &	72.1 &	96.7 \\\hline
\end{tabular}
\caption{\label{tab:all_traffic_categ_perf_complete}Complete performance results on one vs all categorization of different traffic classes and encryption techniques. The task was to classify each class versus all other classes on the test dataset.}
\end{table}

\begin{table}[H]
\centering
\begin{tabular}{|c c c|} \hline
\textbf{Traffic class} & \textbf{Our accuracy(\%)} & \textbf{Flowpic's accuracy(\%)}\\\hline
Unencrypted &    		99.7  &    		97.0\\\hline
Tor &       	    	99.4  &    		85.7\\\hline
VPN &       	    	99.9  &    		99.7\\\hline
\end{tabular}
\caption{\label{tab:encryption_categ_perf_complete}Average accuracies for on one vs all categorization of different traffic classes for each encryption technique. The reported results are the averages of results in Table \ref{tab:perf_comparison_total} for each encryption technique. The corresponding Flowpic accuracies, from Table 3 (rows 4-6) in \cite{Shapira2019}, are always lower.}
\end{table}
% ===================================================

% \section*{References}
\bibliography{bibfile}

% ===================================================

\end{document}